\pdfoutput=1

\documentclass[11pt]{article}

\usepackage[table,xcdraw]{xcolor}

\usepackage[]{ACL2023}

\usepackage{times}
\usepackage{latexsym}
\usepackage{todonotes}
\usepackage{multicol}
\usepackage{lipsum}
\usepackage{multirow}
\usepackage{booktabs}

\usepackage{float}
\usepackage{soul}

\usepackage[T1]{fontenc}

\usepackage[utf8]{inputenc}

\usepackage{microtype}

\usepackage{inconsolata}

\title{LCT-1 at SemEval-2023 Task 10: Pre-training and Multi-task Learning for Sexism Detection and Classification}

\author{Konstantin Chernyshev\footnotemark[1] \qquad Ekaterina Garanina\footnotemark[1] \qquad Duygu Bayram \\ \bf{Qiankun Zheng \qquad Lukas Edman} \\
University of Groningen \\
\texttt{\{k.chernyshev, e.garanina, d.bayram.1, q.zheng.9\}@student.rug.nl} \\
\texttt{j.l.edman@rug.nl}}

\begin{document}
\maketitle
\renewcommand{\thefootnote}{\fnsymbol{footnote}}
\footnotetext[1]{Equal contribution.}
\renewcommand{\thefootnote}{\arabic{footnote}}

\begin{abstract}
Misogyny and sexism are growing problems in social media. Advances have been made in online sexism detection but the systems are often uninterpretable. SemEval-2023 Task 10 on Explainable Detection of Online Sexism aims at increasing explainability of the sexism detection, and our team participated in all the proposed subtasks. Our system is based on further domain-adaptive pre-training \citep{gururangan-etal-2020-dont}. Building on the Transformer-based models with the domain adaptation, we compare fine-tuning with multi-task learning and show that each subtask requires a different system configuration. In our experiments, multi-task learning performs on par with standard fine-tuning for sexism detection and noticeably better for coarse-grained sexism classification, while fine-tuning is preferable for fine-grained classification\footnote{The source code is available at \url{https://github.com/lct-rug-2022/edos-2023}.}.
\end{abstract}

\section{Introduction}

Sexism has been appearing frequently in online spaces in recent years, which not only makes online spaces unfriendly but also exacerbates social prejudice and causes serious harm to targeted groups. In order to control and mitigate it, considerable efforts have been made to detect online sexism \citep{fersini2018overview1, fersini2018overview2, trac2-dataset, Fersini2020, PLN6389}. SemEval-2023 Task 10 on Explainable Detection of Online Sexism (EDOS) \citep{edos2023semeval} aims to improve interpretability via flagging sexist content (Task A) and further deciding on a particular type of sexism in the text, including 4-category (Task B) and 11-category (Task C) classification systems. In this paper, we present our participation in all EDOS subtasks.


Given the complexity of obtaining annotated data, the research on using additional data via different training techniques is quite intensive. \citet{gururangan-etal-2020-dont} illustrated the benefits of further pre-training and introduced two methods for it: only using the task data or only using the domain data. These approaches, used individually and in combination, are shown to perform well in domain-specific tasks with relatively low costs of computing resources.

\citet{safi-samghabadi-etal-2020-aggression} developed a unified end-to-end neural model using a multi-task learning (MTL) approach to address the tasks of aggression and misogyny detection. \citet{lees2020jigsaw} pre-trained a BERT-based model on 1 billion comments and fine-tuned the model with multilingual toxicity data before fine-tuning it on the target dataset. They both demonstrated that using data from similar tasks and fine-tuning the model with it in a multi-task way could improve the model's performance.  

Inspired by the above studies, we explore the impact of further pre-training as well as multi-task learning on the EDOS tasks in our work. Specifically, we collect several datasets that have been developed for the related tasks (Section \ref{section:data}) and include them with various annotation schemes, e.g., binary labeling of hate speech (HS), categorization of target groups, fine-grained misogyny and sexism classification.

First, we run experiments to choose the most suitable preprocessing for our models (Section \ref{section:preprocessing}) including emoji normalization \cite{retrofitting, bornheim-etal-2021-fhac}, hashtag segmentation \cite{NULI}, and masks for contents like usernames and links \cite{paraschiv2019upb, zeinert_annotating_2021}.

Next, we conduct further pre-training on the 2 million texts provided by the organizers and the other hate speech data (Section \ref{section:pretraining}). We experiment with both domain-adaptive and task-adaptive pre-training strategies proposed by \citet{gururangan-etal-2020-dont}. Finally, we investigate whether the model can benefit from additional in-domain data (Section \ref{section:multitask}) using the MTL approach.

The contributions of this paper are as follows:
\begin{enumerate}
    \item We find that further domain-adaptive pre-training can improve the performance on our tasks.
    \item We conclude that the benefits of multi-task learning vary across the tasks. It performs on par with standard fine-tuning on Task A and surpasses it on Task B while being inferior to the performance of the fine-tuning on Task C.
\end{enumerate}

\section{System overview}

We treat all tasks as classification problems. Specifically, we use pre-trained Transformer-based \cite{transformers} models and focus on three approaches to enhance the model performance: data collection and preprocessing, additional pre-training, and MTL.

\subsection{Data}
\label{section:data}

\begin{table}[t]
\resizebox{\linewidth}{!}{
\begin{tabular}{lr|lr}
\toprule
\multicolumn{2}{c|}{\textbf{HS and related}} & \multicolumn{2}{c}{\textbf{Sexism and misogyny}} \\
\midrule
\textbf{Dataset} & \textbf{Entries} & \textbf{Dataset} & \textbf{Entries} \\
\midrule
OLID & 14100 & AMI@EVALITA2018 & 5000 \\
HatEval & 13000 & Call me sexist & 11339 \\
Measuring HS & 39565 & EXIST & 5644 \\
UB & 448000 & Online misogyny & 6355 \\
\bottomrule
\end{tabular}}
\caption{Datasets in English used for pre-training and MTL. All statistics are reported after dataset processing specific for our purposes.}
\label{table:raw_datasets}
\end{table}

The EDOS dataset consists of 14,000 messages collected from Reddit and Gab, the latter social media website known for its politically far-right user base. The messages are annotated into three different levels of granularity. The most general labeling scheme, Task A, is a binary classification task to detect whether a message is sexist. Task B details 4 categories of sexism within the 3,398 \textit{sexist} messages of Task A, which are divided further into 11 categories in Task C. Moreover, the organizers include 2 million non-labeled texts from the same websites.

We split the EDOS dataset 80:20 into train and evaluation sets and use our evaluation set to rank the results of our experiments. Throughout this paper, our evaluation set is titled \textit{eval}, whereas \textit{dev} and \textit{test} refer to the development and test sets of EDOS, respectively.

Apart from the data provided by the organizers, we select\footnote{For selecting the relevant datasets, we used \url{https://hatespeechdata.com} \citep{vidgen2020directions}.} several related datasets annotated for hate speech (HS), sexism, and other related concepts. We divide the datasets into two groups: HS (including related tasks), and sexism or misogyny datasets with further fine-grained annotation. Table \ref{table:raw_datasets} contains statistics for all datasets;  their detailed descriptions, as well as shorter identifiers, are located in Appendix \ref{appendix:datasets}. We have also collected the data in other languages to investigate the influence of multilingual training on the target tasks (Appendix \ref{appendix:multilingual}).

For pre-training on HS data, we use all collected datasets for HS and sexism. For MTL, we recompile the task datasets from the original data; the details are provided in Section \ref{section:multitask}.

\subsection{Preprocessing}
\label{section:preprocessing}

We consider the following preprocessing steps:
\begin{itemize}
    \item Creating a uniform cleaning method across datasets. Since we use several datasets from different authors, the raw data does not always have usernames and URLs masked, and the existing masks differ. We use regular expressions to ensure that all usernames and URLs are masked and that all the masking tokens are the same.
    \item Normalizing hashtags. We use regular expressions to detect hashtags (\#) and apply English word segmentation\footnote{\url{https://pypi.org/project/wordsegment/}}.
    \item Converting emojis to their natural language counterparts with the \texttt{emoji} Python library\footnote{\url{https://pypi.org/project/emoji/}}.
\end{itemize}


\begin{table*}[t]
\centering
\resizebox{\textwidth}{!}{
\begin{tabular}{lllrl}
\toprule
\textbf{Task} & \textbf{Code} & \textbf{Labels} & \textbf{Entries} & \textbf{Datasets} \\
\midrule
Hate speech & hs & binary & 33000 & Measuring HS, HatEval \\
Offensive language & offensive & binary & 14100 & OLID \\
Toxicity & toxic & binary & 40000 & UB \\
Target (within HS) & target & \begin{tabular}[c]{@{}l@{}}\textsc{ind, grp,}\\ \textsc{unt, oth}\end{tabular} & 10110 & HatEval, OLID, AMI@EVALITA2018 \\
Gender mentioned & gender & binary & 79565 & Measuring HS, UB \\
Sexism & sexism & binary & 28338 & all sexism datasets \\
\bottomrule
\end{tabular}}
\caption{Compiled hate speech tasks for MTL for Task A. The \textit{Code} column provides the dataset codes for further reference in MTL discussion.}
\label{table:hs_tasks}
\end{table*}

\subsection{Further pre-training}
\label{section:pretraining}

One line of our research is exploring the further pre-training strategies following \citet{gururangan-etal-2020-dont}. Specifically, we train the existing pre-trained model in an unsupervised manner using Masked Language Modeling (MLM) objective. We consider the following approaches: 1) domain-adaptive pre-training (DAPT): further pre-training of a model using domain-related data available, 2) task-adaptive pre-training (TAPT): utilizing only target task text data in an unsupervised manner, and 3) sequential application of the described techniques (DAPT+TAPT).

As the data for DAPT, we use 2 million texts from Reddit and Gab provided by the organizers and the collected HS data described above. For TAPT, we use the EDOS task data only, including both sexist and non-sexist texts.

\subsection{Multi-task learning}
\label{section:multitask}

To make use of the available annotated HS and sexism data, we apply the multi-task learning approach \cite{caruana_multitask_1997}. This approach assumes training on multiple tasks at once using a single model. Since the advent of BERT models, it is common to use a shared encoder and separate task-specific heads. In this setup, the loss is averaged among the heads. We consider MTL to be beneficial since it indirectly enriches relatively scarce target task data and provides the model with more information about hate speech and sexism.

For MTL we use the MaChAmp \citep{van-der-goot-etal-2021-massive} toolkit. During multi-task learning on multiple datasets, it first splits the datasets into batches (each batch contains instances from one dataset only) and then concatenates and shuffles the split batches before training. During training, losses are averaged with pre-defined weights to represent the final loss. The best model is selected based on the aggregated metric.

We define two sets of tasks that we consider for Task A and Tasks B and C accordingly. MTL for Task A is based on HS datasets since we hypothesize that the variety of HS-related tasks can enhance the capabilities of the model in this domain and thus increase the performance on the target task. HS tasks are presented in Table \ref{table:hs_tasks}. When compiling the task data, Measuring HS and UB datasets were cut with random sampling to avoid heavy imbalance inside the task dataset.


\begin{table}[H]
\centering
\begin{tabular}{llr}
\toprule
\textbf{Task} & \textbf{Code} & \textbf{Entries} \\
\midrule
AMI@EVALITA2018 & evalita & 2245 \\
Call me sexist & sexist & 1241 \\
EXIST & exist & 2794 \\
Online misogyny & online & 448 \\
\bottomrule
\end{tabular}
\caption{Sexism and misogyny classification tasks for MTL for Tasks B and C. The \textit{Code} column provides the dataset codes for further reference in MTL discussion.}
\label{table:sexism_tasks}
\end{table}

Since Tasks B and C are aimed at more precise sexism classification, we apply MTL on sexism datasets with different category systems. We consider each dataset to be a separate task; Table \ref{table:sexism_tasks} contains the statistics per task. For details on classification in each dataset, we refer the reader to the corresponding papers. In addition to the external datasets listed in Table \ref{table:sexism_tasks}, we experiment with adding Task C to Task B MTL and vice versa.


\section{Experiments}

We conduct a series of fine-grained experiments on preprocessing, further pre-training and multi-task learning. We compare several preprocessing components, and the main focus of pre-training and MTL experiments is the input data. We do experiments sequentially, applying the findings from the previous step to the next ones.

\subsection{Evaluation}

Adhering to the official target metric of the shared task, we use the F1-macro score for the intermediate and final evaluation of all models.

While working on the submission version, we primarily used our eval set for evaluating the experiments and made final decisions via online submission to the dev leaderboard; the test set was not available. In this paper, we report all dev and test scores based on the data released by the organizers after the end of the shared task.

\subsection{Baseline}

For the baseline, we considered a variety of state-of-the-art pre-trained models, using the \texttt{base}-sized models. The performance of the best models is shown in Table \ref{table:baseline}. Despite the fact that \textsc{hateBERT} \cite{caselli-etal-2021-hatebert} is slightly better at Task A and \textsc{DeBERTa-v3} \cite{he2021debertav3} has the best performance in Task B, we opted for \textsc{RoBERTa} \cite{roberta} since it has a stable high score over all tasks. For most of 
the subsequent experiments, we used the \textsc{RoBERTa-large} model, which is known to yield better performance. 
\begin{table}[H]
    \resizebox{\linewidth}{!}{
        \centering
        \begin{tabular}{lccc}
            \toprule
            \textbf{Model} & \textbf{Task A} & \textbf{Task B} & \textbf{Task C} \\ 
            \midrule
            \textsc{RoBERTa-base}    & 0.8205    & 0.6034       & \textbf{0.3599}  \\
            \textsc{hateBERT} & \textbf{0.8295}  & 0.6052        & 0.2990    \\
            \textsc{DeBERTa-v3-base} & 0.8088  & \textbf{0.6217} & 0.3192           \\
            \bottomrule
        \end{tabular}
    }
    \caption{F1-macro score for top-3 baseline models on the eval set. The bold font indicates the best result for each task.}
    \label{table:baseline}
\end{table}

\subsection{Preprocessing}

Considering the preprocessing options described in Section \ref{section:preprocessing}, we tested various combinations of components by fine-tuning the \textsc{RoBERTa-base} model. Since usernames and links are not masked in some of the datasets, we do masking regardless of the resulting score, thus focusing on the effect of normalizing emojis and hashtags. 

The results are displayed in Table \ref{table:preprocessing-experiments}. Based on the obtained results, we proceeded only with the unification of masks and normalization of emojis, leaving hashtags intact.


\begin{table}[ht]
    \centering
    \begin{tabular}{cccc}
        \toprule
        \textbf{Masks} & \textbf{Emoji} & \textbf{Hashtags} & \textbf{Task A} \\ 
        \midrule
        + & - & -                         & 0.8110          \\
        + & + & -                         & \textbf{0.8172} \\
        + & - & +                         & 0.8123          \\
        + & + & +                         & 0.8169          \\
        \bottomrule
    \end{tabular}
    \caption{F1-macro eval score of \textsc{RoBERTa-base} model fine-tuned on Task A. The preprocessing is mask normalization, emoji normalization, and hashtags normalization respectively. The bold font indicates the best score.}
    \label{table:preprocessing-experiments}
\end{table}

\subsection{Pre-training}
\label{section:pretraining_exp}

\begin{table*}[t]
    \centering
    \begin{tabular}{l ccc ccc ccc}
        \toprule
        \textbf{Dataset} & \multicolumn{3}{c}{\textbf{Task A}} & \multicolumn{3}{c}{\textbf{Task B}} & \multicolumn{3}{c}{\textbf{Task C}} \\ 
        \cline{2-10} 
         & \textbf{eval} & \textbf{dev} & \textbf{test} & \textbf{eval} & \textbf{dev} & \textbf{test} & \textbf{eval} & \textbf{dev} & \textbf{test} \\ 
        \midrule
        Baseline & 0.8456 & 0.8463 & 0.8514 & 0.6544 & 0.6621 & 0.5997 & 0.5306 & 0.4504 & 0.4621 \\
        \midrule
        EDOS    & 0.8377 & 0.8508  & 0.8517 & 0.6394 & \textbf{0.6723}  & 0.6338   &  0.4975    & 0.4573  & 0.4650      \\
         \cellcolor[HTML]{EBEBEB}2M      & \textbf{0.8474} & 0.8456  & \textbf{0.8534}  & 0.6602           & 0.6422  & 0.6056  &  \cellcolor[HTML]{EBEBEB}0.5374    & \cellcolor[HTML]{EBEBEB}\textbf{0.5201}  & \cellcolor[HTML]{EBEBEB}0.4764          \\
        2M+HS & 0.8408 & \textbf{0.8612}  & 0.8476  &  \textbf{0.6756} & 0.6655  & \textbf{0.6359}   &  \textbf{0.5463}  & 0.4713  & \textbf{0.4908}     \\
        \bottomrule
    \end{tabular}
    \caption{Further MLM pre-training of \textsc{RoBERTa-large} model using TAPT and DAPT approaches, with fine-tuning on tasks A, B, and C. The models are scored with F1-macro. Best scores for each task and set are in bold, submitted model is highlighted with grey.}
    \label{table:pretraining}
\end{table*}

We used the pre-trained \textsc{RoBERTa-large} model for our experiments. We further pre-train the models using MLM task until convergence of validation loss, which we define as non-decreasing loss for 5 consecutive evaluation steps. The parameters are fixed across experiments (Appendix \ref{appendix:pretraining_params}).

We conducted the training on EDOS data only (TAPT), and on 2M texts alone or concatenated with the collected HS data (DAPT). We have also attempted sequential training on an extended dataset and EDOS data (DAPT+TAPT), but it did not perform well in our preliminary experiments. We trained the models using only mask normalization as preprocessing, as it proved to yield better results. The resulting language models were fine-tuned and tested on the target tasks. The final f1-macro scores are shown in Table \ref{table:pretraining}.

For further experiments, including MTL, we used the obtained \textsc{RoBERTa-large} model pre-trained on 2M texts for Task A and C, and \textsc{RoBERTa-large} model pre-trained on 2M+HS texts for Task B. Models for Tasks A and B were selected based on eval score, while the model for Task C was selected by its dev score due to its exceptionally high value.

\subsection{Multi-task learning}

\begin{table*}[t]
\centering
\begin{tabular}{llccc}
\toprule
\textbf{Task} & \textbf{Datasets} & \textbf{eval} & \textbf{dev} & \textbf{test} \\
\midrule
\multirow{3}{*}{A} & \cellcolor[HTML]{EBEBEB}offensive + FT & \cellcolor[HTML]{EBEBEB}\textbf{0.8524} & \cellcolor[HTML]{EBEBEB}0.8521 & \cellcolor[HTML]{EBEBEB}0.8446 \\
 & offensive, target & 0.8363 & \textbf{0.8553} & 0.8470 \\
 & hs + FT & 0.8460 & 0.8539 & \textbf{0.8585} \\
 \midrule
\multirow{2}{*}{B} & \cellcolor[HTML]{EBEBEB}edosC, evalita + FT & \cellcolor[HTML]{EBEBEB}\textbf{0.6777} & \cellcolor[HTML]{EBEBEB}0.7108 & \cellcolor[HTML]{EBEBEB}0.6277 \\
 & edosC, evalita, exist & 0.6463 & \textbf{0.7236} & \textbf{0.6575} \\
\midrule
\multirow{2}{*}{C} & edosB, sexist + FT & \textbf{0.5489} & 0.4515 & \textbf{0.4854} \\
 & edosB, sexist & 0.5155 & \textbf{0.4892} & 0.4518 \\
\bottomrule
\end{tabular}
\caption{F1-macro scores for best MTL models on eval, dev and test sets. Base models: \textsc{RoBERTa-large} further pre-trained on 2M for Tasks A and C, 2M+HS for Task B. +FT: further fine-tuning on the target task after MTL.  Best scores are in bold, submission systems are highlighted with grey. We did not submit a MTL model for task C due to its inferior performance on the dev set compared to standard fine-tuning.}
\label{table:multitask}
\end{table*}

Using the MaChAmp toolkit, we fixed the training parameters for all tasks (Appendix \ref{appendix:multitask_params}), used pre-trained models selected in the previous step (Section \ref{section:pretraining_exp}), and experimented on the dataset combinations. All combinations included the corresponding EDOS dataset.

As opposed to testing all dataset combinations, we did the experiments incrementally. First, we tested all datasets separately, i.e. MTL on each dataset paired with the target EDOS dataset. Afterwards, we formed triples and subsequently larger sets from the most prominent options. For the best dataset combinations based on our eval set, we conducted further fine-tuning on the target task only. 

The results of MTL experiments are presented in Table \ref{table:multitask}. Moreover, we conducted several experiments on using multilingual data and models and observed a major drop in the performance. We discuss these experiments in Appendix \ref{appendix:multilingual}.

\section{Discussion}

For two approaches that we consider -- fine-tuning and MTL -- we generally selected the best models in each setup relying on eval scores and made the final choice between the two setups using the online development set. Evaluation of all our models on now available dev and test sets reveals the ranking mismatch among the partitions. This indicates both the complexity of the task and the partition unevenness, both of which greatly complicate the best model selection. Nevertheless, we can observe certain trends.

Our pre-training experiments show that the concept of further pre-training can be beneficial for the downstream task performance of sexism detection and classification. Different input data performed better for different tasks, although TAPT achieves lower scores compared to DAPT. Although such pre-training requires additional computational resources, the resulting language model can potentially be reused for other downstream tasks. 

Considering MTL, the model for Task A benefits more from the tasks of a target, offensive language, and binary HS classification, which are only somewhat related to sexism. It can be due to the fact that the model gains a larger variety of information from more distant tasks. Performance on Tasks B and C is improved by joint training on these two tasks. Nevertheless, the effect of adding other datasets appears arbitrary, making it difficult to draw any conclusions. Another inconsistency is the impact of further fine-tuning, which varies quite significantly among the experiments. 

Due to the above-mentioned partition unevenness, we observed the major score decrease of the submitted models on the test set, with the alternative system variations performing better by a large margin. The final test scores of the submitted models are 0.8446 for Task A, 0.6277 for Task B, and 0.4764 for Task C. 

Going beyond submitted models, further domain adaptation pre-training improves the quality of sexism detection and classification compared to the baseline, and the selection of the further fine-tuning method depends greatly on the task. MTL outperforms standard fine-tuning for Task B and shows comparable results on Task A. For Task C, the results are consistently in favor of standard fine-tuning with carefully chosen hyperparameters.


\section{Conclusion}

Explainable models for online sexism detection are important for building a safe environment and mitigating interpretability problems. However, our results show that high performance becomes more difficult to achieve for such a complicated task as the labels become detailed and the task becomes fine-grained. 

We found that domain-adaptive further pre-training of a language model improves its performance on a downstream task. Built on the domain-adapted models, MTL and standard fine-tuning behave differently depending on the task, which means that the task formulation has a heavy impact on the model selection even in the case of in-domain data.

\section*{Acknowledgements}

We would like to thank Tommaso Caselli for the insightful supervision and multiple wonderful ideas.

We would like to thank the Center for Information Technology of the University of Groningen for their support and for providing access to the Peregrine high-performance computing cluster.

The first four authors are supported by the Erasmus Mundus Master's Programme ``Language and Communication Technologies''\footnote{\url{https://lct-master.org}}. 

\bibliography{anthology,custom}
\bibliographystyle{acl_natbib}

\appendix

\section{Datasets}
\label{appendix:datasets}

Here we list all datasets that we collected and used either in the main set of experiments (Table \ref{table:raw_datasets}) or for exploring multilingual MTL (Table \ref{table:multilingual_datasets}).

The hate speech (HS) datasets used include:

\begin{enumerate}
    \setlength\itemsep{0em}
    \item Offensive Language Identification Dataset (OLID) \citep{zampieri-etal-2019-predicting} -- offensive language in tweets.

    \item HatEval \citep{basile-etal-2019-semeval} -- HS against immigrants and women in Twitter in English and Spanish.

    \item Hate Speech Detection (HaSpeeDe) \citep{bosco2018overview, manuela2020haspeede} -- HS in Italian social media, including Twitter, Facebook, and news.

    \item Measuring Hate Speech (Measuring HS) \citep{kennedy2020constructing} -- fine-grained annotation of HS (including aggregated severity score) and target identity groups in social media.

    \item Jigsaw Unintended Bias in Toxicity Classification (UB) \citep{borkan2019nuanced} -- annotation for toxicity and target identity groups on the CivilComments platform. We worked with examples with annotated identities, which is roughly a quarter of the whole dataset.

\end{enumerate}

The datasets on sexism and misogyny include:

\begin{enumerate}
    \setlength\itemsep{0em}
    \item Automatic Misogyny Identification at EVALITA 2018 evaluation campaign (AMI@EVALITA2018) \citep{fersini2018overview1} -- misogyny in tweets in Italian and English.

    \item Call me sexist, but... (Call me sexist) \citep{samory2021sexism} -- fine-grained sexism annotation on English social media data done by multiple workers. For our purposes, we use only messages where the class can be derived from the absolute majority of votes among the workers.

    \item sEXism Identification in Social neTworks 2021 (EXIST) \citep{PLN6389} -- sexism identification and classification in tweets in English and Spanish.

    \item The Expert Annotated Online Misogyny Dataset (Online misogyny) \citep{guest-etal-2021-expert} -- misogyny classification of substrings in the text. In our work, we use full texts and consider only texts with all spans belonging to one category.

\end{enumerate}

\section{Further pre-training parameters}
\label{appendix:pretraining_params}


Most of the pre-training parameters follow \citet{gururangan-etal-2020-dont}. The parameters we updated are (TAPT / DAPT respectively):
\begin{itemize}
    \setlength\itemsep{0em}
    \item masking probability: 15\%;
    \item batch size: 32 / 24;
    \item maximum number of epochs: 10 / 5;
    \item learning rate: 5e-6 for both pre-training and further fine-tuning.
\end{itemize}

\section{Multi-task learning parameters}
\label{appendix:multitask_params}

We kept the original task dataset sizes, applied equal loss weights for all tasks, and made the following changes to the default MaChAmp v0.4 training parameters:
\begin{itemize}
    \setlength\itemsep{0em}
    \item batch size: 4;
    \item no discriminative fine-tuning and gradual layer unfreezing;
    \item learning rate: 5e-6 for multi-task and 1e-6 for further fine-tuning.
\end{itemize}

We did MTL for 20 epochs and further fine-tuning for 10 epochs.

\section{Multilingual MTL}
\label{appendix:multilingual}

For exploring the multilingual MTL, we used multilingual (e.g. EXIST) and entirely non-English datasets (e.g. HaSpeeDe). We conducted two experiments: MTL on Task A with multilingual version of \textit{hs} dataset, which contains texts in English, Italian, and Spanish, and MTL on Task B with \textit{evalita} and \textit{exist} datasets, also comprised of these three languages. The multilingual dataset statistics are shown in Table \ref{table:multilingual_datasets}. We used \texttt{roberta-large} model for English-only setup and \texttt{xlm-roberta-large} for the multilingual one.

\begin{table}[t]
\resizebox{\linewidth}{!}{
\begin{tabular}{llrl}
\toprule
\textbf{Task} & \textbf{Lang} & \textbf{Entries} & \textbf{Datasets} \\
\midrule
\multirow{3}{*}{\textit{hs}} & en & 33000 & HatEval, Measuring HS \\
 & es & 6600 & HatEval \\
 & it & 12600 & HaSpeeDe \\
 \midrule
\multirow{2}{*}{\textit{evalita}} & en & 2245 & \multirow{2}{*}{AMI@EVALITA2018} \\
 & it & 2337 &  \\
 \midrule
\multirow{2}{*}{\textit{exist}} & en & 2794 & \multirow{2}{*}{EXIST} \\
 & es & 2864 & \\
\bottomrule
\end{tabular}}
\caption{Task datasets for multilingual MTL. The codes en, es, it stand for English, Spanish, and Italian respectively.}
\label{table:multilingual_datasets}
\end{table}

\begin{table}[ht]
    \centering
    \begin{tabular}{lrr}
    \toprule
    \textbf{} & \textbf{A, hs} & \textbf{B, evalita, exist} \\ 
    \midrule
    Multilingual & 0.8175 & 0.5963 \\
    English & \textbf{0.8382} & \textbf{0.6259} \\ 
    \bottomrule
    \end{tabular}
    \caption{F1-macro scores of models trained on multilingual and English-only data in the MTL setup.}
    \label{table:multilingual}
\end{table}

The results are presented in Table \ref{table:multilingual}. Since the multilingual setup performs noticeably worse, we decided not to pursue this direction of research.

\end{document}